\documentclass[letterpaper, 10 pt, conference]{ieeeconf}  

\IEEEoverridecommandlockouts                              

\usepackage{amsmath} 
\usepackage{amssymb}  
\usepackage{tikz}
\usetikzlibrary{positioning, arrows.meta, quotes}
\usetikzlibrary{shapes,snakes}
\usetikzlibrary{bayesnet}
\tikzset{>=latex}
\tikzstyle{plate caption} = [caption, node distance=0, inner sep=0pt,
below left=5pt and 0pt of #1.south]
\usepackage{algorithmic}
\usepackage{graphicx}
\usepackage{subcaption}
\usepackage{changepage}
\usepackage{textcomp}
\usepackage{xcolor}
\usepackage{url}
\usepackage{multirow}
\usepackage[bookmarks=true]{hyperref} 
\usepackage{array}
\usepackage{booktabs}
\usepackage{tabularray}
\usepackage{makecell}
\usepackage{times}
\usepackage{comment}
\usepackage{cite}
\usepackage{balance}

\title{Design and Evaluation of a Collaborative Teleoperation Interface \\ for Multi-Operator Multi-Robot Systems in Challenging Environments}

\author{Alperen Kenan$^{1}$ $^{\star}$, Juan José García Cárdenas$^{2}$ $^{\star}$, Paul Bremner$^{1}$, Adriana Tapus$^{2}$ and Manuel Giuliani$^{3}$
\thanks{$^{\star}$These authors contributed equally to this work.}
\thanks{$^{\dagger}$This work was supported by the European Commission's Marie Skłodowska-Curie Actions (MSCA) Project RAICAM (GA 101072634), and UK Research and Innovation (UKRI) grant number EP/X025977/1.}
\thanks{For the purpose of open access, the author has applied a Creative Commons Attribution (CC BY) licence to any Author Accepted Manuscript version arising.}
\thanks{$^{1}$Alperen Kenan is a PhD candidate and Paul Bremner is an Associate Professor in Human-Robot Interaction at the Bristol Robotics Laboratory, University of the West of England, Bristol, United Kingdom
        {\tt\small alperen.kenan@uwe.ac.uk, paul2.bremner@uwe.ac.uk}}
\thanks{$^{2}$Juan José García Cárdenas is a PhD candidate and Adriana Tapus is Full Professor in the Computer Science and System Engineering Department (U2IS), Autonomous Systems and Robotics Lab, ENSTA, Institut Polytechnique de Paris, Paris, France
        {\tt\small juan-jose.garcia@ensta.fr; adriana.tapus@ensta.fr}}
\thanks{$^{3}$Manuel Giuliani is a Professor at Kempten University of Applied Sciences, Kempten, Germany
        {\tt\small manuel.giuliani@hs-kempten.de}}}

\begin{document}

\maketitle
\thispagestyle{empty}
\pagestyle{empty}

\begin{abstract}

Multi-robot systems that combine the complementary capabilities of specialised platforms are critical for complex missions in challenging, inaccessible environments. Current deployments assign an operator or operator team to each robot, with collaboration relying on operator-to-operator communication to manually coordinate heterogeneous platforms with incompatible interfaces. This operator-dependent approach increases cognitive load and limits scalability across robots from different manufacturers. This study designs and evaluates a shared-visualisation collaborative teleoperation interface to improve operator coordination, situational awareness, and the task performance in multi-operator, multi-robot systems. 

A comparative user study with 16 participants (8 participant pairs) compared three teleoperation conditions: co-located operators with a shared interface, remote operators with a shared interface, and co-located operators with individual interfaces. Participants controlled two simulated robots, a sensor-equipped drone and a manipulator equipped ground robot, in a nuclear power plant digital twin to complete disaster response tasks: radiation source identification and deactivation, and locating and retrieving a rescue dummy to safety. Performance metrics, physiological data, and behavioural measures were recorded to comprehensively assess task performance, cognitive load, and operator awareness across conditions. The co-located shared-interface condition achieved the fastest mission completion ($292.8$~s), outperforming remote operation ($351.6$~s) and individual interfaces ($385.1$~s), with the fewest collisions ($0.9$ per trial) and highest situational awareness accuracy ($91.1\%$). Physical co-presence was associated with reduced cognitive load, with NASA-TLX scores increasing from $38.2$ in co-located to $57.7$ in remote operation, supported by consistent physiological indicators. These results demonstrate that co-presence and shared interfaces provide complementary benefits, supporting workload regulation and mutual awareness respectively in multi-operator, multi-robot teleoperation. 

This research presents a collaborative teleoperation interface that reduces operator cognitive load and improves task performance in multi-robot systems, addressing critical gaps in operator communication and robot coordination for safety-critical remote operations.

\end{abstract}

\section{Introduction} 

Hazardous and inaccessible environments, such as nuclear power plants (NPPs), present significant risks to humans during maintenance, decontamination, and emergency response tasks \cite{iaea2016}. Robotic teleoperation has emerged as a key strategy to mitigate these risks, reducing human exposure to radiation while maintaining operational capability where human presence is unsafe \cite{sheridan1992telerobotics, vitanov2021suite}. However, the complexity of such missions often exceeds the capabilities of a single platform, driving the adoption of multi-robot systems that combine the complementary strengths of specialised robots to meet the demands of increasingly challenging operations.

Existing research on multi-robot systems often assumes a single operator controlling multiple robots, while real-life critical missions typically assign a dedicated operator or team to each robot, with different teams collaborating to achieve a common goal \cite{kenan2025robot}. In such deployments, coordination relies on communication between operators to manually synchronise heterogeneous platforms with incompatible interfaces, increasing workload and limiting scalability. Without a unified interface that supports shared situational awareness and coordinated decision-making, teams must interpret data streams independently, increasing the risk of miscommunication and mission failure in safety-critical environments.

To address this gap, a collaborative teleoperation interface for multi-operator, multi-robot systems in safety-critical environments is presented, in which operators share a unified visualisation of the remote environment and all robots while retaining individual control of their own robot. The interface is validated through a comparative user study with 16 participants. It is designed to: (i) support shared situational awareness by integrating multi-robot camera and status feeds in a unified display; (ii) improve coordination and task performance by enabling real-time information sharing between operators and providing intuitive coordinated control; and (iii) enhance operator safety by supporting remote operation in hazardous environments.

\begin{figure}[t]
\centering
\includegraphics[width=1.00 \columnwidth]{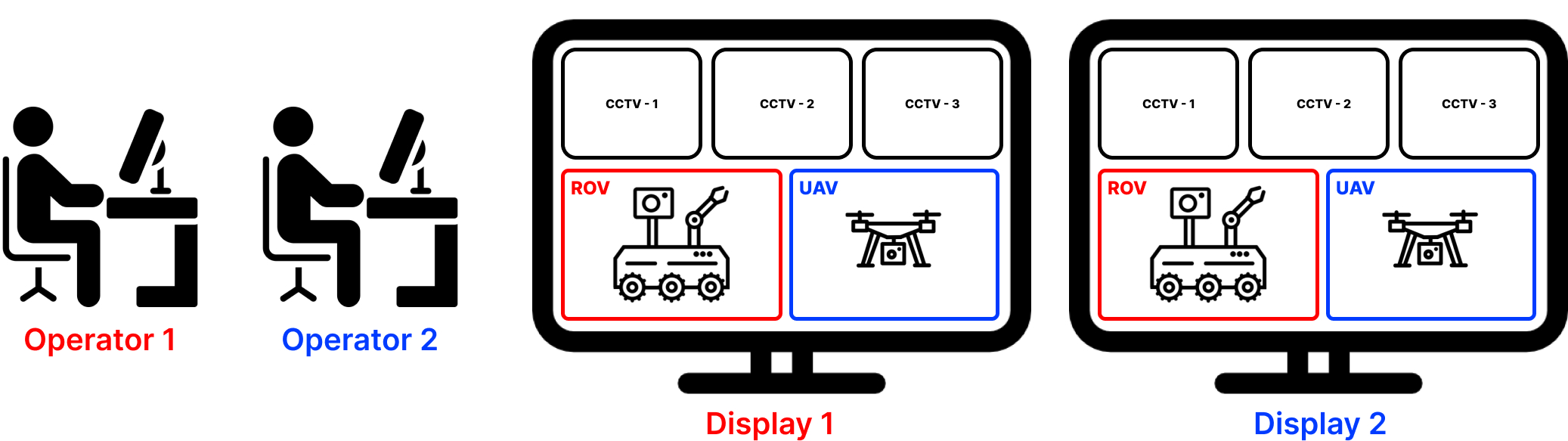}
\caption{Illustration of the collaborative teleoperation interface}
\label{fig:illustration}
\vspace{-15pt}
\end{figure}

Figure \ref{fig:illustration} illustrates the working principle of the proposed interface, in which multiple specialised robots are deployed in a remote environment inaccessible to humans, while co-located operators coordinate through a unified interface from a safely located control room, maintaining shared awareness of both the robots and their surrounding environment.

This paper makes three primary contributions: (1) the design of a collaborative teleoperation interface for multi-operator multi-robot systems in challenging environments; (2) an open-source simulation environment to support researchers studying robotics and collaborative interfaces in safety-critical settings; and (3) a user study with 16 participants providing experimental evidence on the effectiveness of collaborative teleoperation interfaces.

The remainder of the paper is organized as follows. Section \ref{sec:relatedwork} reviews prior work on multi-robot teleoperation and cognitive load estimation. Section \ref{sec:Methodology} outlines the experimental design, including participants, conditions, and data collection. Section \ref{sec:simulation} details the simulation environment and disaster response task scenarios. Section \ref{sec:Hypothesis} presents the study hypotheses. Section \ref{sec:results} reports performance, workload, and physiological results. Section \ref{sec:discussion} discusses the findings and limitations, and Section \ref{sec:Conclusions} summarizes the main findings and their implications for collaborative teleoperation interface design in multi-operator multi-robot systems.

\section{Related Work}
\label{sec:relatedwork}

Robotic teleoperation in challenging environments, such as nuclear, biological and industrial settings, has played a key role in reducing human exposure to risk while ensuring task precision \cite{sheridan1992telerobotics, vitanov2021suite}. In nuclear environments specifically, robots must contend with high radiation levels, contamination, access limitations, and communication constraints \cite{decommissioning_challenges}. The Fukushima Daiichi disaster exemplified these challenges, exposing critical limitations of single-robot systems under the scale and unpredictability of real-world emergency response \cite{nagatani2013emergency, murphy2016disaster}.These scenarios highlight the need for multi-robot systems that distribute tasks across specialised platforms, driving a research shift toward coordinated multi-robot teams for tasks such as search and rescue \cite{darmanin2017review}. To support realistic operator training and evaluation, high-fidelity digital twins of real nuclear facilities have been developed, enabling immersive multi-robot teleoperation research grounded in real-world conditions \cite{baniqued2024multimodal}. In this study, an open-source digital twin of a real nuclear power plant is developed from data collected on-site \cite{tian2025low}, incorporating task scenarios, user interfaces, mobile robots, and associated controller scripts for sensing and manipulation, made publicly available for other researchers to build upon.

Managing multi-robot coordination in high-consequence settings requires careful interface design to ensure precise and cognitively efficient teleoperation. Effective interfaces must balance information density with operator cognitive load, incorporating structured guidelines for visual layout, state awareness, and scalability \cite{adamides2014usability}. Restricted perception in remote operations makes situational awareness particularly critical, with user studies identifying key requirements across environmental feedback, system status, and operator workload \cite{graf2020user}. Ultimately, effective collaboration in multi-operator, multi-robot systems demands scalable shared-control frameworks that unify distributed data into an intuitive operational representation. 

Multi-robot systems require robust strategies for task allocation, communication, and decentralised control \cite{yan2013survey, cortes2017coordinated}, with added complexity when coordinating heterogeneous platforms such as drones and ground robots \cite{rizk2019cooperative}. The introduction of multiple human operators further increases communication and control challenges. Early work addressed low-level stabilisation of multi-operator teleoperation \cite{sirouspour2005multi}, while more recent approaches optimise network topologies for coordination \cite{tianmoroco} and incorporate AR/MR interfaces and digital twins to reduce operator workload \cite{martins2024multi, adekoya2026multi}.

A critical but under-explored challenge in multi-operator systems is the cost of communication between operators. Grounding theory highlights the importance of co-presence in reducing coordination overhead, as co-located operators can rely on implicit coordination through gaze, gesture, and shared attention, while remote operators must make coordination explicit through verbal communication, increasing cognitive burden and reducing awareness of team member activity \cite{clark1991grounding, riley2008performance}. While distributing control across an operator team might reduce workload per operator, it introduces coordination overhead and the risk of conflicting commands \cite{kolling2015human}. Shared interfaces have been shown to improve shared cognition and implicit coordination, whereas individual interfaces require operators to verbally reconstruct a shared mission picture before joint decisions can be made \cite{lowry2013signal}. Structured control-sharing methodologies that align human reasoning with robotic precision are therefore needed to manage this coordination cost \cite{music2017control}.

In contrast to work on autonomous control, network optimisation, and hardware-intensive multi-robot systems, this work examines how screen-based shared visualisation and operator co-location affect performance and team dynamics. Shared co-located, shared remote, and individual configurations are compared in a high-fidelity nuclear disaster-response simulation, combining performance metrics, questionnaires, and physiological measures.

Trust and cognitive load are critical factors in HRI, influencing performance and willingness to use an interface. Physiological sensing has emerged as a powerful method for estimating these variables in real time. Shared-control systems have been shown to reduce workload and moderately enhance trust via pupil diameter and NASA-TLX \cite{Pan2024}, while multimodal signals such as EDA, heart rate variability, and blink dynamics enable high-resolution cognitive load profiling \cite{Wang2024}. GSR alone has achieved 69\% accuracy in detecting trust during safety-critical handovers \cite{Campagna2024}, and cross-modal sensing combining EEG, behavioural, and physiological signals further improves estimation fidelity in teleoperation contexts \cite{Yang2024, Zakeri2023}. This study combines blink rate, GSR, and thermal imaging with behavioural and performance metrics.

\section{Methodology}
\label{sec:Methodology}

To evaluate the proposed collaborative interface for robot teleoperation, a comparative user study was conducted under three different conditions (Fig.~\ref{fig:conditions}). The objective was to assess usability, cognitive load, and task performance. Tasks were modelled on realistic disaster response scenarios such as radiation detection and search and rescue, which required collaboration between operator pairs. Details of the simulation and the tasks are explained in Sec \ref{sec:simulation}.


\subsection{Participants}

A total of 16 participants (8 participant pairs) took part in 
the study, 10 male and 6 female. Participant ages ranged from 23 to 45 years (\emph{M} = 32.8, \emph{SD} = 3.2). The group was geographically diverse, with 9 participants from Europe, 2 from Asia, 2 from Africa, and 3 from South America. Educational backgrounds included 5 PhD students, 8 Master students, and 3 undergraduates. Most participants reported some prior experience with robots (13 out of 16) and considered themselves proficient with controllers or simulators (11 out of 16). All expressed a positive outlook on the growing role of robots in daily life and workplace tasks.

\subsection{Study Environment and Context}

The study took place at the Autonomous Systems and Robotics Lab, ENSTA, Institut Polytechnique de Paris, Paris, France where participant pairs were located during the user studies. Before the experiment sessions, participants completed a pre-experiment questionnaire covering age, gender, academic background, prior robotics exposure (laboratory courses, DIY projects, commercial cobots), and weekly hours spent on video gaming or other joystick-based activities, in line with similar experimental setups \cite{Cheng2024}. Each session started with a study introduction, followed by a brief tutorial and a familiarization period, allowing participants to gain confidence before proceeding.

\begin{figure}[ht]
\centering
\includegraphics[width=1.00 \columnwidth]{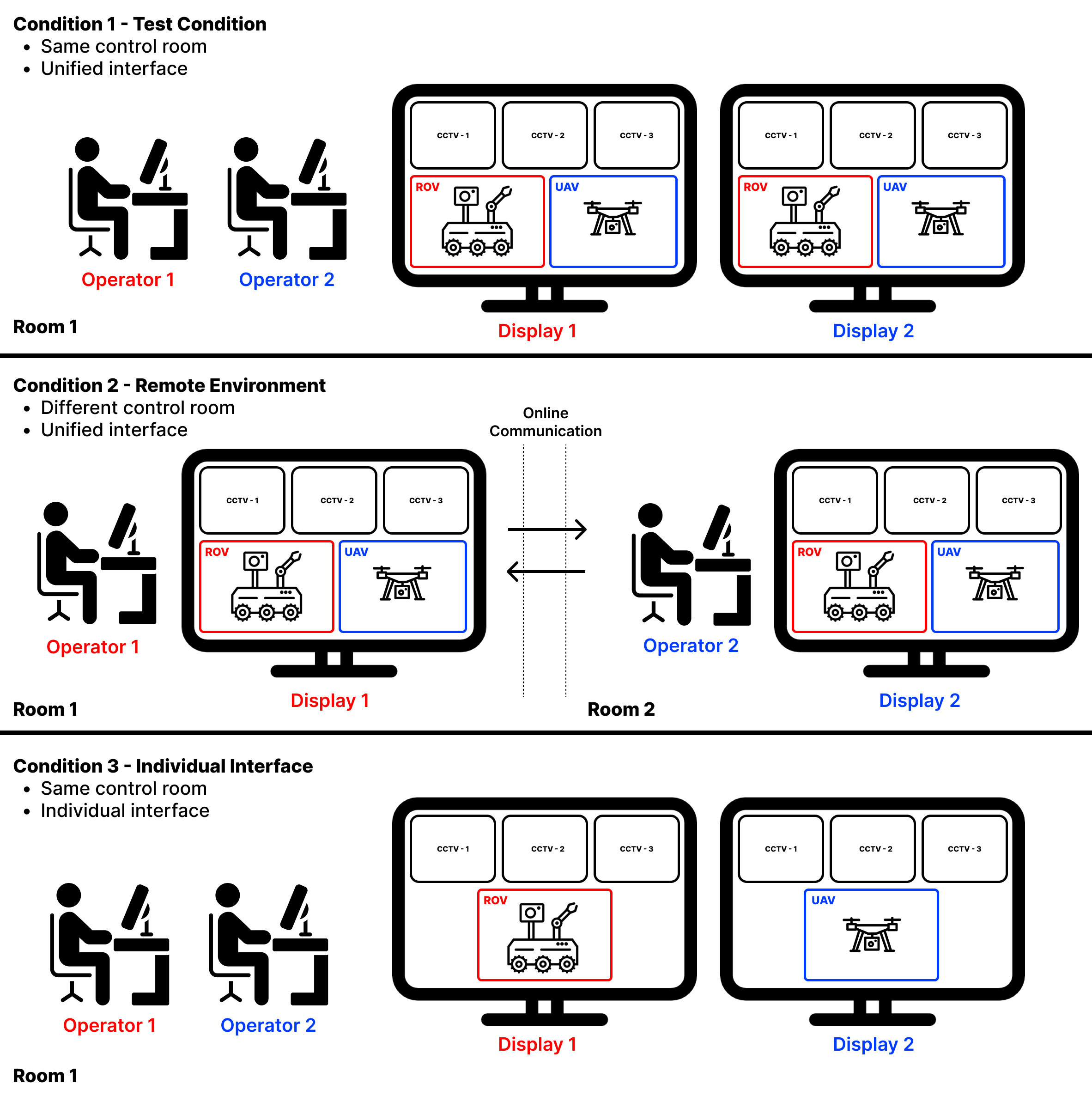}
\caption{Three conditions used during the user studies}
\label{fig:conditions}
\vspace{-15pt}
\end{figure}

Teleoperation tasks were performed under three conditions and Fig. \ref{fig:conditions} shows these conditions used during the user studies: (C1) co-located operators with a shared interface, (C2) remote operators with a shared interface, and (C3) co-located operators with individual interfaces. Further details of the user interfaces are provided in Section \ref{sec:ui_cases}.All control conditions used the same main computer for the simulation while the display was in separate monitors, to isolate computing and delay effects of the simulation. Joystick controller and keyboard were standard commercial products. User interface details are given in Section~\ref{sec:ui_cases}.

8 participant pairs each executed the tasks described in Sec \ref{sec:simulation} in all three control modes. Multiple task performance metrics were logged automatically by the simulation, including task completion time, number of collisions per robot, number of false detections of the radiation source, and the distance between the ground robot and the drone for task performance evaluation.

In addition to task metrics physiological signals and cognitive activity were recorded for each user, which will be explained in more detail in the subsection below. The order of conditions was randomized between participants to minimize learning effects. To reduce potential bias, participants were not informed of the study’s hypotheses. The open-source code for this project is publicly available. \footnote{The repository is available at: \url{https://github.com/kenanalperen/NPP-Digital-Twin}. It contains the \texttt{.unitypackage} file with the digital twin environment, task scenarios, mobile robots, and the associated controller scripts for sensing and manipulation.}

\subsection{Cognitive Load and Trust Evaluation}

To assess cognitive workload and operator trust, both qualitative and quantitative data were collected. After each condition, participants completed the NASA-TLX \cite{nasatlx} for six subjective workload sub-scores, and following their final run, provided open-ended feedback on usability, and trust. These were supported by physiological measures including blink rate, facial temperature, and galvanic skin response (GSR), which reflect cognitive workload and stress \cite{garcia2025legged} \cite{garcia2024}, together providing a multimodal record of operator workload alongside performance metrics.

High-resolution physiological and behavioural signals were captured via RGB video, thermal imaging, and electrodermal sensing. Facial features were recorded with a Logitech C930e webcam (1920×1080 px, 30 fps) \cite{logitechC930e} and the OpenFace ROS node for AU45 extraction, alongside an Optris PI-640 thermal camera (640×480 px, 32 fps) \cite{optrisPI640}; thermal frames were mapped to 68 dlib facial landmarks to extract mean temperatures in the nose, periorbital, and forehead regions \cite{dlib}. Skin conductance was monitored with Shimmer 3 GSR+ units \cite{shimmerGSR3} on the index and middle fingers of the dominant hand, streaming at 128 Hz to capture phasic responses tracking moment-to-moment cognitive demands \cite{boucsein2012electrodermal}. Because GSR is non-invasive and unaffected by respiratory artifacts, it offers a reliable proxy for sympathetic arousal without the encumbrance of cardio-respiratory belts \cite{berntson2007cardiovascular}. All sensor feeds were integrated into a custom teleoperation dashboard displaying live video, thermal ROIs, GSR traces, and blink counts, enabling online cognitive-load monitoring and event annotation. This unified pipeline ensured cross-modal synchronisation and provided the ground truth for the adaptive workload modelling in subsequent sections.

\subsection{Ethics}

This study was approved by the Research Ethics Committee of the University of the West of England, College of Arts, Technology \& Environment (Reference: 15480310). 

A participant information sheet was provided to all participants prior to their consent to take part, and signed consent forms were collected before the experiments started. Appropriate measures were taken to ensure participant confidentiality and data security throughout the study. Participants had the right to withdraw at any point and to request the removal of their data up to seven days following the study.

\section{Digital Twin Simulation Implementation}
\label{sec:simulation}

\begin{figure}[ht]
\centering
\includegraphics[width=1.00\columnwidth]{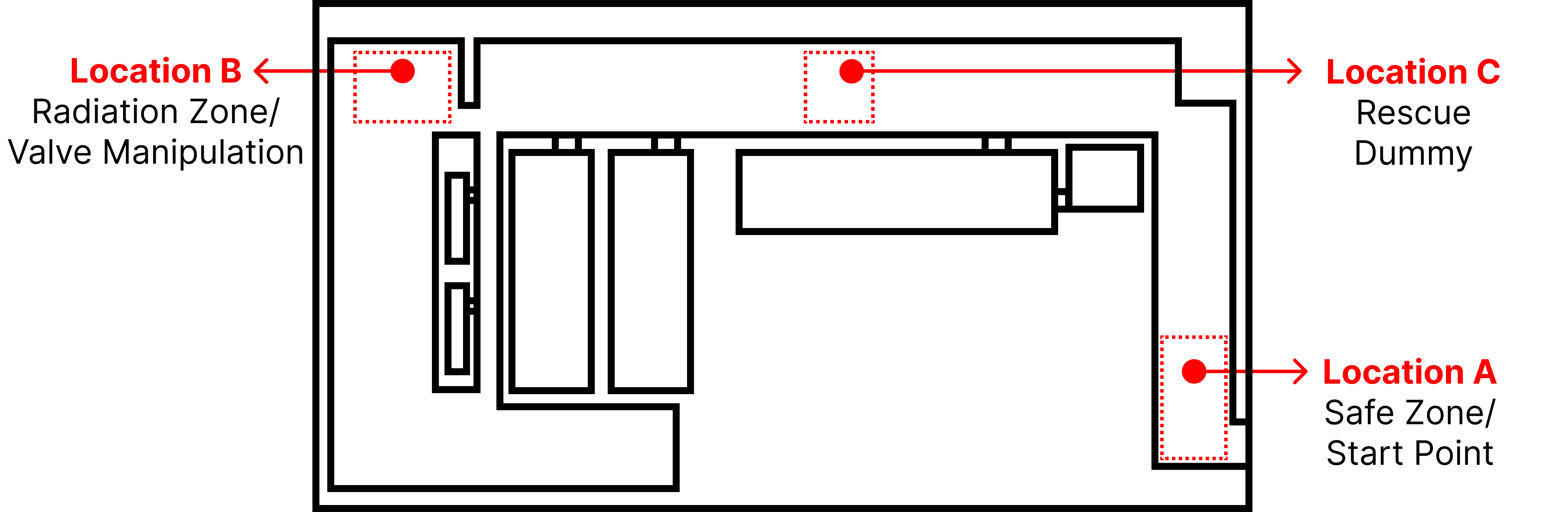}
\caption{Ground-Floor Map of the Nuclear Plant.}
\label{fig:ground_map}
\vspace{-15pt}
\end{figure}

 The environment is a digital twin of the Zwentendorf Nuclear Power Plant (NPP) in Austria, developed by the authors as part of the RAICAM team during the 5th European Robotics Hackathon using video and point cloud data. For the current research, the digital twin was extended with mobile robots, robot control and interaction algorithms, a collaborative User Interface (UI), and task scenarios. Further details on the hackathon participation and digital twin creation process can be found in \cite{tian2025low}. The user study environment and mobile robots were simulated using Unity (version 6.4)\footnote{Unity is a real-time 3D development platform used for simulation, visualisation, and interactive applications. See \url{https://unity.com/}.}. Fig. \ref{fig:ground_map} displays the ground-floor map of the NPP. Location A serves as both the starting point of the mobile robots and the safe zone for retrieving the rescue dummy. Location B contains the radiation source and the valve used to deactivate it. Location C contains the rescue dummy to be retrieved to the safe zone. Although the second floor of the NPP, including the engine room, is present in the environment, it was not used in the task scenario.

\subsection{Mobile Robots}

\begin{figure}[ht]
\centering
\includegraphics[width=1.00\columnwidth]{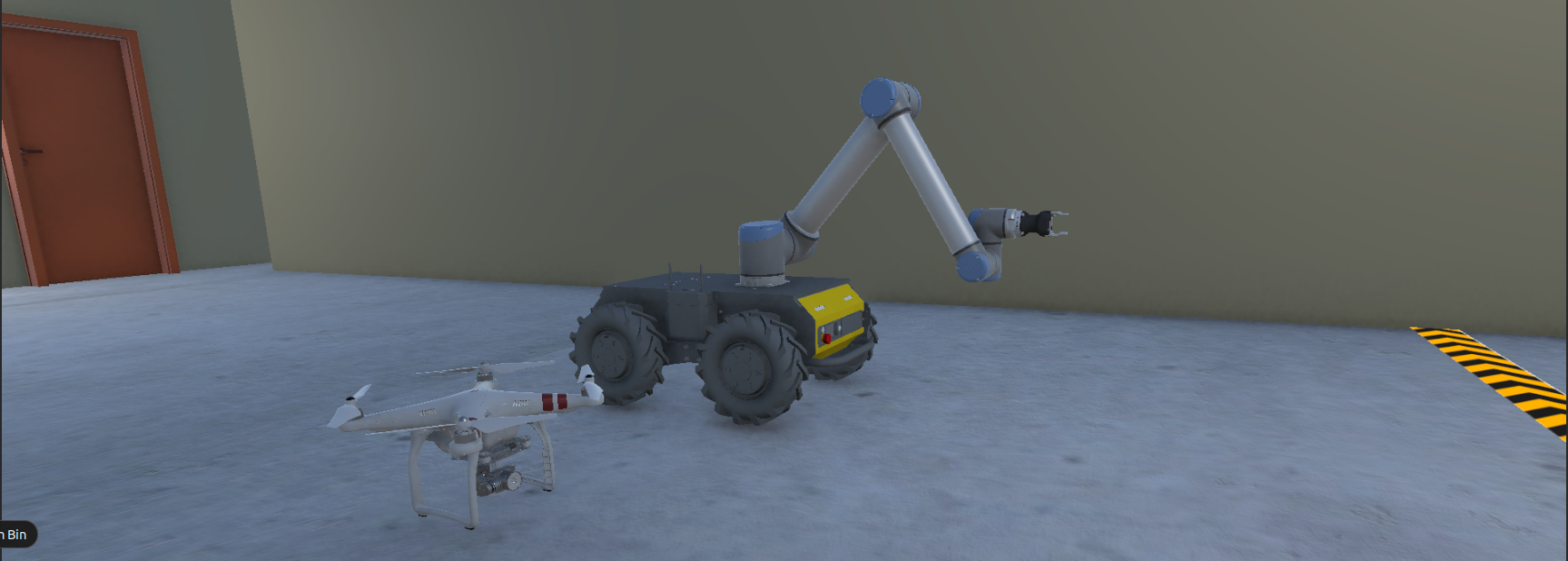}
\caption{Mobile robots in initial positions inside the safe zone.}
\label{fig:mobile_robots}
\vspace{-13pt}
\end{figure}

Two mobile robots were implemented in the task scenario based on real-world CAD models:\footnote{CAD models were based on the following platforms: Clearpath HUSKY A300: \url{https://clearpathrobotics.com/husky-unmanned-ground-vehicle-robot/}; Universal Robots UR10: \url{https://www.universal-robots.com/products/ur10-robot/}; DJI Phantom 3: \url{https://www.dji.com/phantom-3-pro}. The open-source UR10 robot model for Unity simulation is available at: \url{https://github.com/Preliy/Flange}.} (i) a ground mobile robot (Clearpath HUSKY A300) with a Universal Robots UR10 manipulator and a ROBOTIQ Hand attached, and (ii) a drone (DJI Phantom 3) equipped with a radiation sensor and a body camera with a controllable angle. Manipulation of the valve and grasping of the rescue dummy were simplified to a single click when the manipulator was within range of the target object. Both robots are displayed in Fig. \ref{fig:mobile_robots}.

Both robots were controlled by separate operators using different interfaces: the ground robot and manipulator via keyboard, the drone via joystick. Keyboard and joystick were chosen as widely available industrial input devices, deliberately different across roles to reflect heterogeneous manufacturer controllers.

The drone had a limited battery, and restricted communication range requiring it to remain in close proximity to the ground robot, and no manipulator to interact with the environment. The ground robot lacked a radiation sensor and had a restricted field of view from its onboard camera, making collaboration essential for task completion. Role constraints reflect conditions observed in on-site trials \cite{tian2025low}.

\subsection{Disaster Scenario}

\begin{figure}[ht]
    \centering
    \begin{subfigure}[b]{0.48\columnwidth}
        \centering
        \includegraphics[width=\linewidth]{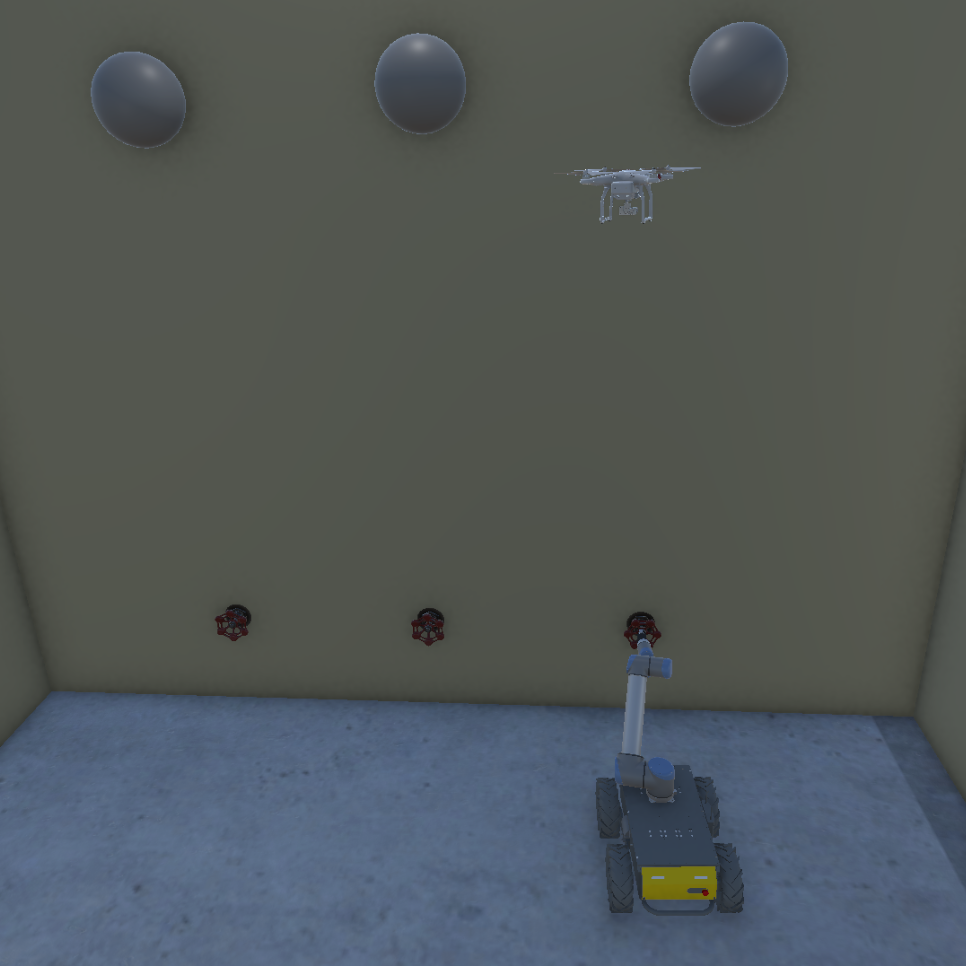}
        \caption{Radiation source detection and valve manipulation.}
        \label{fig:radiation_subfig}
    \end{subfigure}
    \hfill
    \begin{subfigure}[b]{0.48\columnwidth}
        \centering
        \includegraphics[width=\linewidth]{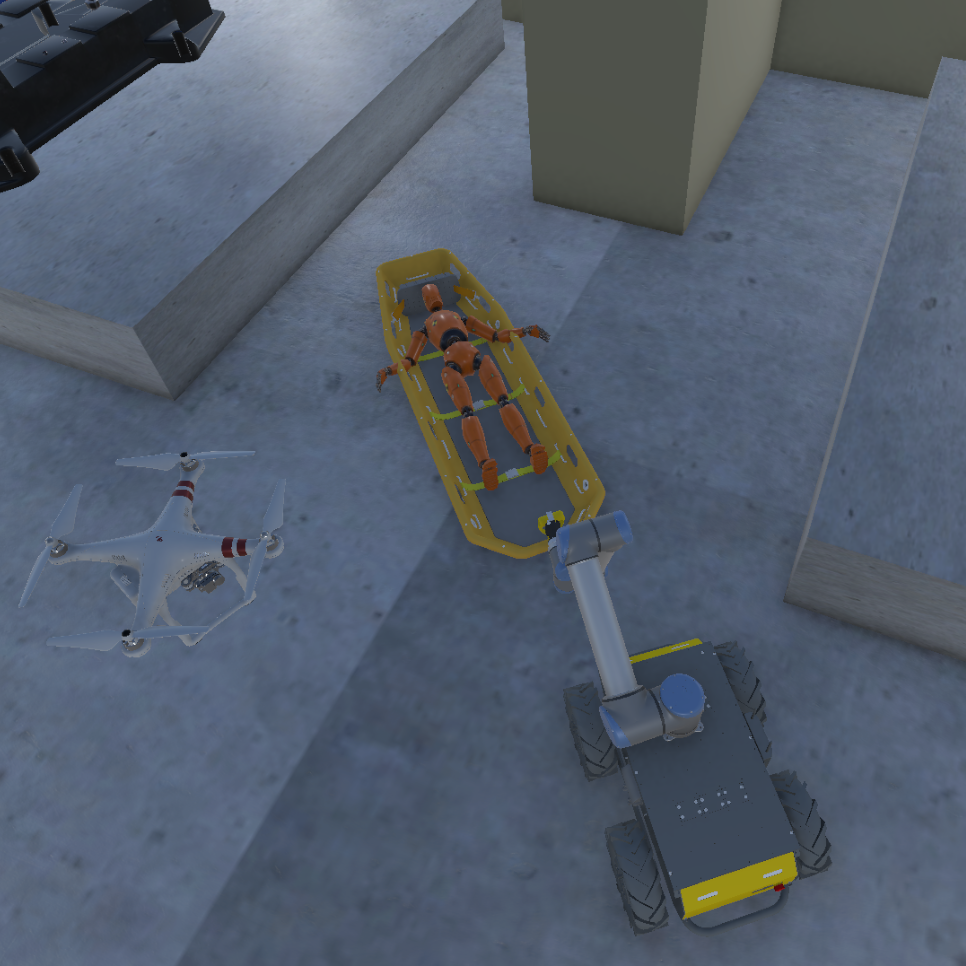}
        \caption{Rescue dummy search and retrieval.}
        \label{fig:dummy_subfig}
    \end{subfigure}
    \caption{Disaster response task scenario components.}
    \label{fig:disaster_scenario}
    \vspace{-10pt}
\end{figure}

The task scenario required operators to collaboratively complete a simulated nuclear disaster response mission using the ground robot and drone. Fig. \ref{fig:disaster_scenario} illustrates the two critical task components: radiation source detection with valve manipulation, and rescue dummy search and retrieval.

In the first phase, operators navigated both robots from Location A to Location B, where the drone identified the active radiation source (randomised per session) and the ground robot deactivated it via valve manipulation. Operators then navigated to Location C, where the ground robot retrieved the rescue dummy and returned it to the safe zone at Location A, with the drone providing visual guidance throughout. Task completion required both the radiation source to be deactivated and the dummy to be safely retrieved.

\subsection{Collaborative User Interface}
\label{sec:ui_cases}

The interface design follows requirements gathered from nuclear-sector end users \cite{kenan2025robot}: information should be accessible to the entire team, job-specific status should be presented without distraction, and views should extend beyond a single camera feed.

\begin{figure}[ht]
\centering
\includegraphics[width=1.00\columnwidth]{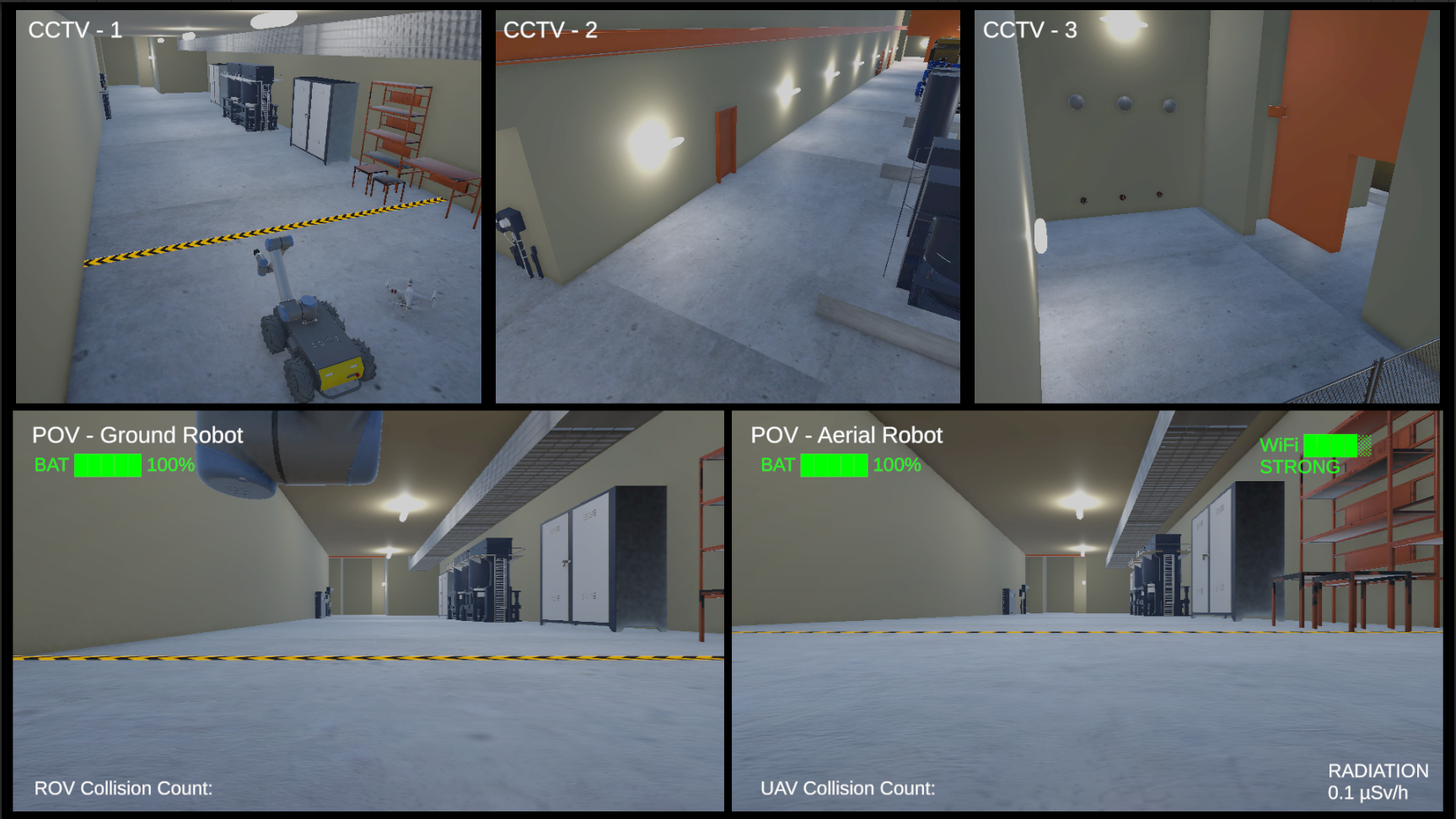}
\caption{Collaborative UI used during the shared interface conditions.}
\label{fig:unified_ui}
\vspace{-15pt}
\end{figure}

Fig. \ref{fig:unified_ui} shows the collaborative UI used during the user studies. The top three screens display CCTV camera views of Locations A, B, and C respectively (from left to right). The two screens below show the body camera feeds from the ground robot (left) and the drone (right). Both feeds display the robot name, remaining battery percentage, and number of collisions. The drone feed additionally displays communication signal strength and radiation measurements.

\begin{figure}[ht]
    \centering
    \begin{subfigure}[b]{0.48\columnwidth}
        \centering
        \includegraphics[width=\linewidth]{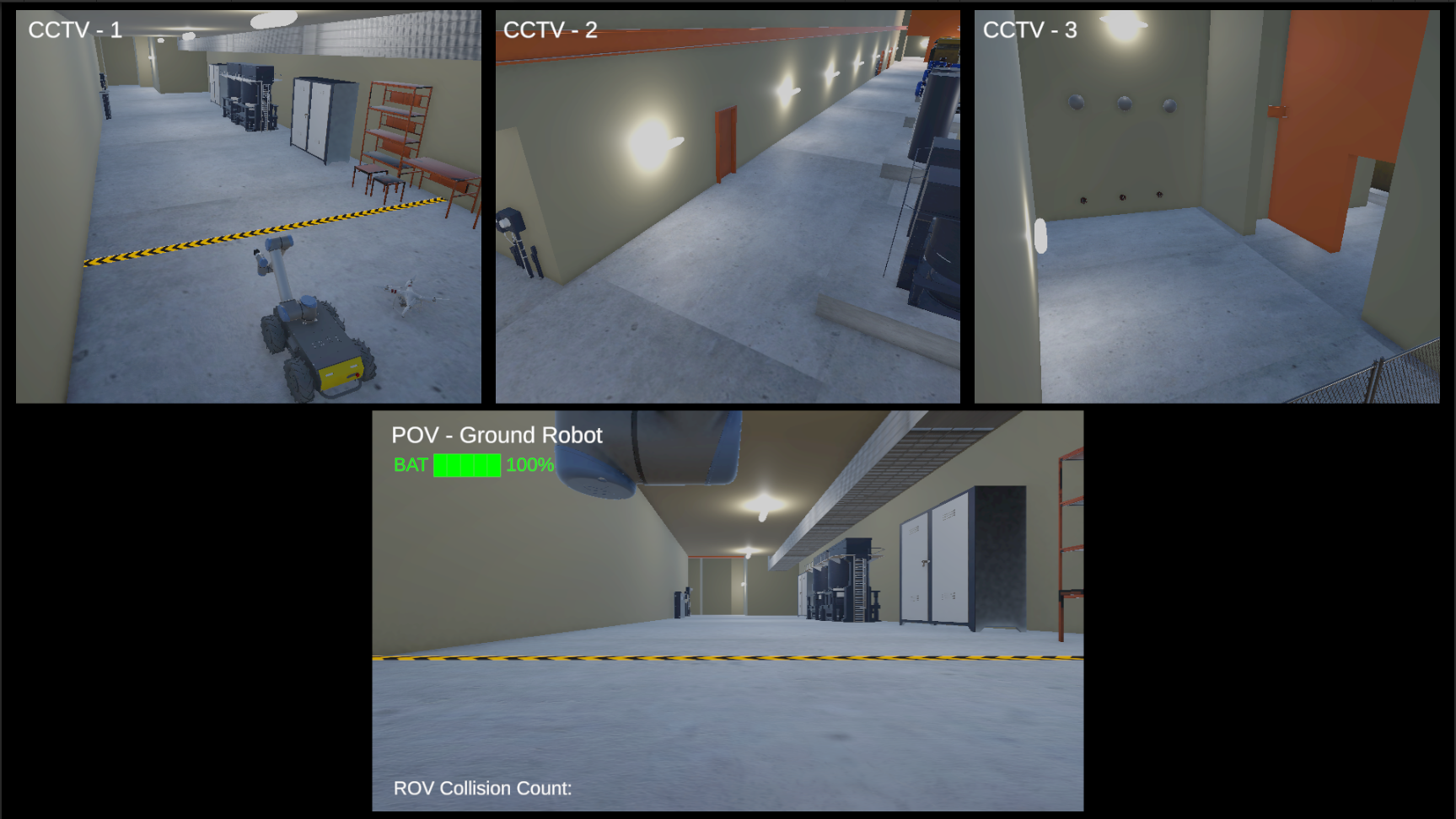}
        \caption{UI of the ROV operator.}
        \label{fig:ground_ui_subfig}
    \end{subfigure}
    \hfill
    \begin{subfigure}[b]{0.48\columnwidth}
        \centering
        \includegraphics[width=\linewidth]{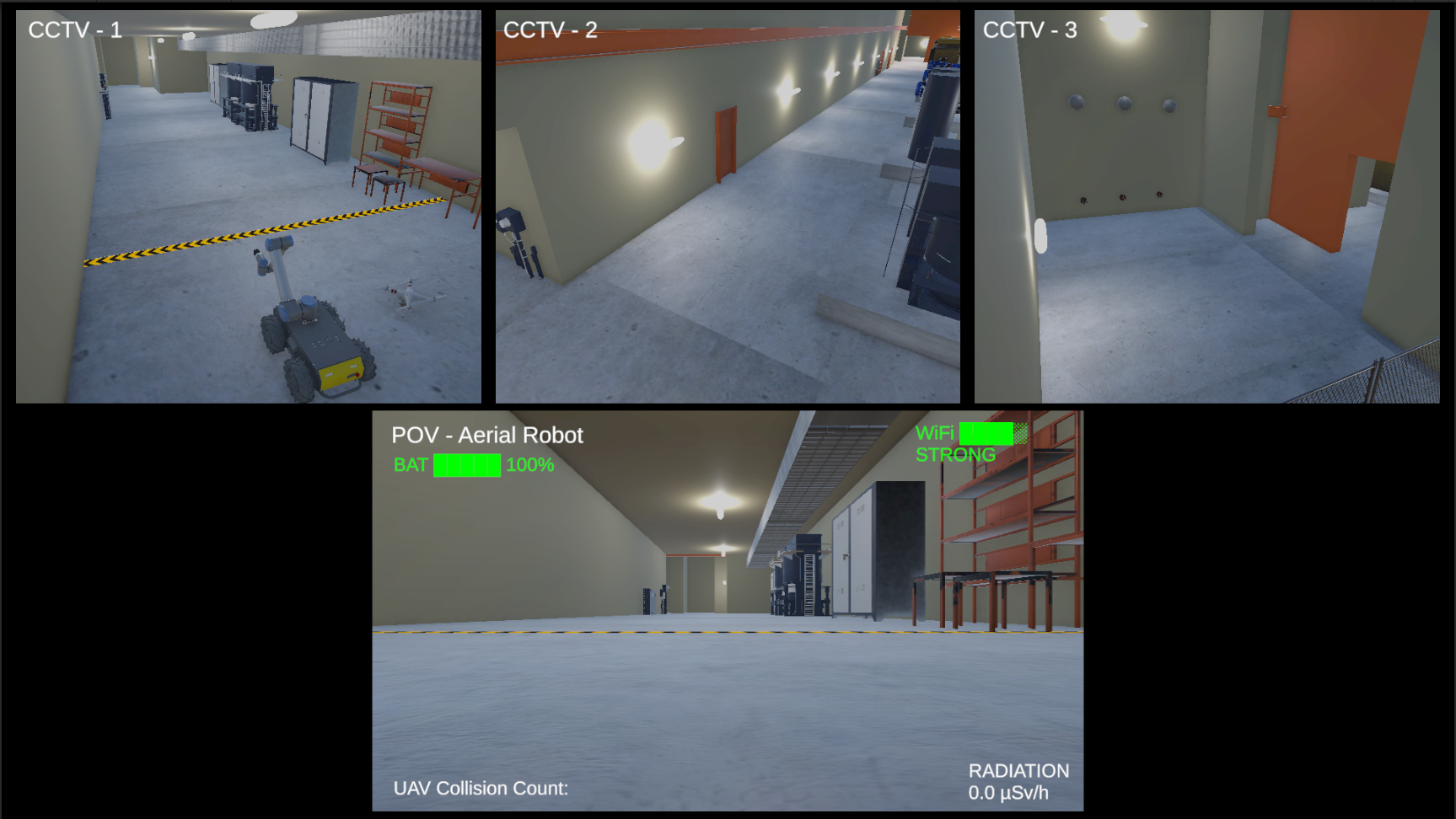}
        \caption{UI of the UAV operator.}
        \label{fig:drone_ui_subfig}
    \end{subfigure}
    \caption{Individual UIs used on separate displays during C3.}
    \label{fig:individual_ui}
    \vspace{-10pt}
\end{figure}

Fig. \ref{fig:individual_ui} displays the individual UIs shown on separate monitors during C3, with the left screen assigned to the ground robot operator and the right to the drone operator. Operators had access only to their own robot's camera feed, could not see their partner's screen, and relied on verbal communication to maintain awareness of the overall situation.

\section{Hypotheses}
\label{sec:Hypothesis}

Building on research in collaborative HRI, three hypotheses are formulated:

\begin{itemize}

    \item [\textbf{H1}] Participants in condition C1 (co-located, shared interface) will have lower cognitive load and better performance than condition C2 (remote, shared interface).
    
    \item [\textbf{H2}] Participants in condition C1 (co-located, shared interface) will have lower cognitive load and better performance than condition C3 (co-located, individual interfaces).
    
    \item [\textbf{H3}] Participants with higher spatial skill or gaming experience will demonstrate lower mental workload and better performance across all three conditions compared to low-skill users.
    
\end{itemize}

Each hypothesis is grounded in prior research in HRI and human factors, as outlined below:

\textbf{H1} is supported by grounding in communication \cite{clark1991grounding}, which highlights the importance of copresence in reducing coordination costs and enabling efficient collaboration. In C1, copresence allows operators to rely on implicit coordination through gaze, gesture, and shared attention. In C2, the loss of copresence forces coordination to become explicit via verbal communication, increasing coordination overhead and reducing awareness of the team member activity \cite{riley2008performance}. 

\textbf{H2} is grounded in research on shared situation awareness, with findings suggesting that shared interfaces improve shared cognition and implicit coordination \cite{lowry2013signal}, whereas individual interfaces require explicit communication and increase coordination effort. In C3, operators must verbally reconstruct a shared mission picture before joint decisions, while in C1 shared access to both vehicle feeds enables team awareness without direct communication.

\textbf{H3} is based on evidence that spatial orientation skill and gaming experience improve efficiency in perspective-taking tasks critical for teleoperation, leading to fewer collisions, lower NASA-TLX ratings, and faster task completion \cite{hegarty2004dissociation, chen2012supervisory}. High-skill participants are expected to show smaller GSR spikes, lower blink-rate increases, and shorter completion times across all conditions compared to low-skill users.

\section{Results}
\label{sec:results}

All eight operator pairs completed the three experimental conditions. Unless otherwise stated, condition effects were first tested using one-way repeated-measures ANOVAs with condition as the within-subject factor. For task-performance and communication outcomes, the operator pair was treated as the unit of analysis; for questionnaire and physiological outcomes, the operator was treated as the unit of analysis. Mauchly's tests were non-significant for all analysed measures, indicating that the sphericity assumption was tenable; therefore, uncorrected degrees of freedom are reported. Significant omnibus effects were followed by planned Holm-corrected pairwise comparisons aligned with the hypotheses: C1--C2 tested the effect of co-presence, whereas C1--C3 tested the effect of the shared interface. Omnibus effect sizes are reported as generalized eta squared ($\eta_G^2$), and pairwise effect sizes as Cohen's $d_z$.

\subsection{Task Performance}
\label{subsec:task_performance}

Task performance differed significantly across conditions. Repeated-measures ANOVAs showed significant omnibus effects of condition for completion time, $F(2,14)=36.13$, $p<.001$, $\eta_G^2=.644$; collision count, $F(2,14)=6.63$, $p=.009$, $\eta_G^2=.360$; handoff latency, $F(2,14)=49.13$, $p<.001$, $\eta_G^2=.689$; and situational-awareness freeze-probe accuracy, $F(2,14)=39.23$, $p<.001$, $\eta_G^2=.632$. These effects indicate that the experimental condition had a large and operationally meaningful impact on mission execution.

Planned comparisons showed that pairs completed the mission faster in C1 ($292.8$~s) than in C2 ($351.6$~s; $d_z=1.67$, $p_{\mathrm{Holm}}=.002$) and C3 ($385.1$~s; $d_z=3.55$, $p_{\mathrm{Holm}}<.001$). Thus, C1 reduced completion time by approximately $59$~s relative to remote operation and by approximately $92$~s relative to the individual-interface condition. Pairs also produced fewer collisions in C1 ($0.9$ collisions/trial) than in C2 ($3.0$ collisions/trial; $d_z=2.06$, $p_{\mathrm{Holm}}=.010$) and C3 ($3.3$ collisions/trial; $d_z=1.47$, $p_{\mathrm{Holm}}=.028$). Handoff latency was lowest in C1 ($6.6$~s), increased in C2 ($11.8$~s; $d_z=1.54$, $p_{\mathrm{Holm}}=.012$), and was highest in C3 ($17.3$~s; $d_z=3.76$, $p_{\mathrm{Holm}}<.001$). Situational-awareness accuracy followed the inverse pattern, with higher accuracy in C1 ($91.1\%$) than in C2 ($80.0\%$; $d_z=1.72$, $p_{\mathrm{Holm}}=.002$) and C3 ($69.3\%$; $d_z=2.84$, $p_{\mathrm{Holm}}<.001$).

\begin{figure*}[t]
    \centering
    \includegraphics[width=0.79\textwidth]{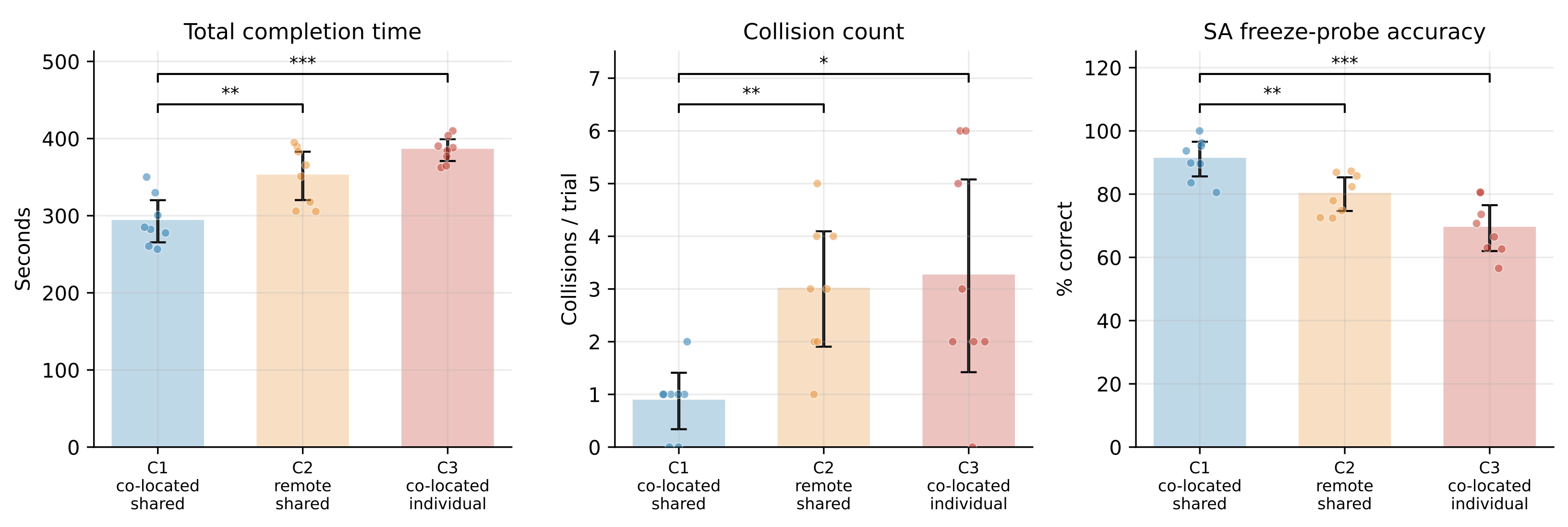}
    \caption{Task performance across the three experimental conditions: mission completion time, collision count, and situational-awareness freeze-probe accuracy. Bars show means, error bars indicate pointwise 95\% confidence intervals, and dots represent operator pairs ($n=8$ per condition). Brackets indicate the planned C1--C2 and C1--C3 comparisons. $^{*}p_{\mathrm{Holm}}<.05$, $^{**}p_{\mathrm{Holm}}<.01$, $^{***}p_{\mathrm{Holm}}<.001$.}
    \label{fig:task_performance}
\end{figure*}

Figure~\ref{fig:task_performance} summarizes the differences in completion time, collision count, and situational-awareness accuracy. C1 consistently showed better performance than C2 and C3 across these outcomes. Together with the handoff-latency results, these findings support both H1 and H2 for task performance: co-presence was beneficial when operators used the shared interface, and interface sharing was beneficial when operators were physically co-located.

\subsection{Cognitive Load and Trust}
\label{subsec:cognitive_load_trust}

Cognitive load showed a significant omnibus effect of condition. NASA-TLX scores differed across conditions, $F(2,30)=48.77$, $p<.001$, $\eta_G^2=.518$, indicating a large condition effect on subjective workload. Physiological workload indicators showed convergent effects, including tonic GSR, $F(2,30)=22.89$, $p<.001$, $\eta_G^2=.338$; heart rate, $F(2,30)=18.81$, $p<.001$, $\eta_G^2=.312$; HRV-RMSSD, $F(2,30)=14.56$, $p<.001$, $\eta_G^2=.227$; and blink-rate elevation, $F(2,30)=13.40$, $p<.001$, $\eta_G^2=.336$.

Planned comparisons showed that workload increased mainly when physical co-presence was removed. NASA-TLX scores were lower in C1 ($38.2$) than in C2 ($57.7$; $d_z=2.19$, $p_{\mathrm{Holm}}<.001$), but C1 did not differ significantly from C3 ($42.0$; $d_z=0.47$, $p_{\mathrm{Holm}}=.266$). The same pattern appeared in the physiological measures. Tonic GSR increased from C1 ($6.5~\mu$S) to C2 ($8.2~\mu$S; $d_z=1.55$, $p_{\mathrm{Holm}}<.001$), while the C1--C3 comparison was not significant (C3: $6.9~\mu$S; $d_z=0.45$, $p_{\mathrm{Holm}}=.218$). Heart rate, HRV-RMSSD, and blink-rate elevation followed the same direction, with significant C1--C2 differences but non-significant C1--C3 differences.

Trust also differed significantly across conditions. Overall trust showed a large omnibus effect, $F(2,30)=71.72$, $p<.001$, $\eta_G^2=.385$, with similar effects for performance trust, $F(2,30)=30.79$, $p<.001$, $\eta_G^2=.311$, and moral trust, $F(2,30)=34.71$, $p<.001$, $\eta_G^2=.370$. Planned comparisons showed that overall trust was higher in C1 ($5.90$) than in C2 ($4.88$; $d_z=1.85$, $p_{\mathrm{Holm}}<.001$), while C1 did not differ significantly from C3 ($5.74$; $d_z=0.28$, $p_{\mathrm{Holm}}=.333$). Performance trust and moral trust followed the same pattern.

Together, these results support H1 for cognitive load and trust, but not H2. Removing co-presence increased workload, physiological arousal, and reduced trust. In contrast, replacing the shared interface with individual interfaces did not significantly affect these measures when operators remained physically co-located.

\subsection{Team Understanding and Situational Awareness}
\label{subsec:team_understanding}

Team-understanding measures showed significant omnibus effects of condition. The Networked-Minds co-presence subscale differed strongly across conditions, $F(2,30)=78.85$, $p<.001$, $\eta_G^2=.701$. Significant effects were also observed for mutual awareness, $F(2,30)=67.35$, $p<.001$, $\eta_G^2=.693$, and mutual understanding, $F(2,30)=36.93$, $p<.001$, $\eta_G^2=.412$.

Planned comparisons revealed a dissociation between co-presence and interface sharing. The co-presence subscale followed the physical arrangement of the operators: scores were high in C1 ($6.09$), decreased in C2 ($3.96$; $d_z=2.96$, $p_{\mathrm{Holm}}<.001$), and remained high in C3 ($6.03$; $d_z=0.09$, $p_{\mathrm{Holm}}=.765$). Thus, perceived co-presence was mainly determined by physical co-location.

By contrast, mutual awareness followed the interface configuration. Mutual awareness was high in C1 ($5.96$), remained relatively high in C2 ($5.46$; $d_z=0.62$, $p_{\mathrm{Holm}}=.130$), and decreased substantially in C3 ($3.53$; $d_z=3.64$, $p_{\mathrm{Holm}}<.001$). Mutual understanding showed the same pattern: C1 ($5.81$) and C2 ($5.51$) were comparable, whereas C3 was significantly lower ($4.19$; $d_z=1.89$, $p_{\mathrm{Holm}}<.001$).

These results indicate that co-location and shared displays support different aspects of teamwork. Physical co-location supports the perceived presence of the partner, whereas the shared interface supports mutual awareness and understanding of the joint task state.

\subsection{Communication Behaviour and Operator-Skill Covariate}
\label{subsec:communication_skill}

Communication behaviour differed significantly across conditions. Repeated-measures ANOVAs showed significant omnibus effects for communication word count, $F(2,14)=59.70$, $p<.001$, $\eta_G^2=.808$; verbal exchanges, $F(2,14)=40.42$, $p<.001$, $\eta_G^2=.724$; clarification requests, $F(2,14)=45.01$, $p<.001$, $\eta_G^2=.649$; and confirmation statements, $F(2,14)=38.45$, $p<.001$, $\eta_G^2=.564$.

Planned comparisons showed that communication volume was lowest when operators were both co-located and using the shared interface. Pairs used an average of $158$ words per trial in C1, compared with $250$ words in C2 ($d_z=2.75$, $p_{\mathrm{Holm}}=.002$) and $299$ words in C3 ($d_z=4.70$, $p_{\mathrm{Holm}}<.001$). Verbal exchanges, clarification requests, and confirmation statements followed the same pattern, increasing significantly when either physical co-presence or shared visual context was removed. This suggests that C1 enabled more implicit coordination, whereas C2 and C3 required more explicit verbal grounding.

Finally, spatial/gaming skill was examined as a covariate across conditions. Higher operator skill was associated with fewer collisions ($r=-.81$, $p=.014$), lower GSR phasic peaks ($r=-.69$, $p=.003$), and fewer GSR spike counts ($r=-.61$, $p=.013$). However, skill did not significantly predict NASA-TLX scores ($r=-.16$, $p=.55$), blink-rate elevation ($r=-.08$, $p=.77$), or completion time ($r=-.38$, $p=.36$). Therefore, H3 was partially supported: higher-skill operators showed fewer control errors and lower electrodermal arousal, but not significantly lower subjective workload or shorter completion times.

\section{Discussion}
\label{sec:discussion}

H1 was supported. The comparison between C1 and C2 shows that physical co-presence had a clear positive effect on collaborative teleoperation. When operators were co-located and used the shared interface, they completed the task faster, produced fewer collisions, showed shorter handoff latencies, and achieved higher situational-awareness accuracy than when they used the same shared interface remotely. This suggests that co-presence reduces coordination overhead even when the visual information available to the operators is similar. The cognitive-load and trust measures further reinforce this interpretation: NASA-TLX, GSR, heart rate, HRV-RMSSD, and blink-rate elevation all indicated higher workload and physiological arousal in C2, while overall trust, performance trust, and moral trust decreased under remote operation. Therefore, the loss of physical co-presence appears to increase the need for explicit verbal grounding and reduce the efficiency of team coordination. In the context of multi-operator multi-robot teleoperation, these results suggest that co-presence supports not only communication efficiency, but also workload regulation, trust, and perceived partnership between operators.

H2 was partially supported. The comparison between C1 and C3 shows that the shared interface improved task performance and team-level awareness, but did not significantly reduce cognitive load or increase trust. Operators using the shared interface completed the mission faster, had fewer collisions, shorter handoff latencies, and higher situational-awareness accuracy than operators using individual interfaces. Mutual awareness and mutual understanding were also substantially higher in the shared-interface conditions, indicating that the unified display helped operators maintain a common representation of the mission state and understand the actions of the other operator and robot. However, NASA-TLX, physiological workload indicators, and trust did not differ significantly between C1 and C3. This indicates that the shared interface mainly contributed to coordination and shared situational awareness rather than directly reducing individual workload. Since operators in C3 were still physically co-located, they could preserve trust and perceived co-presence through direct communication, even though their reduced access to shared visual information impaired coordination. Thus, the shared interface should be interpreted as a mechanism for improving mutual awareness, handoffs, and task performance, rather than as a primary mechanism for reducing cognitive load.

H3 was also partially supported. Higher spatial or gaming skill was associated with fewer collisions, lower GSR phasic peaks, and fewer GSR spike counts, suggesting that skilled operators were better able to control the robots and experienced lower electrodermal arousal during the task. However, skill did not significantly predict NASA-TLX scores, blink-rate elevation, or completion time. This indicates that individual skill can reduce control errors and physiological reactivity, but it does not necessarily make the complete collaborative mission faster or subjectively easier. In this task, performance depended not only on the ability of each operator to control their own robot, but also on team-level factors such as handoff timing, shared awareness, communication efficiency, and coordination between heterogeneous robots. Therefore, training and prior experience may help reduce local control difficulty, but they cannot fully compensate for weak shared awareness or interface limitations. Overall, the results show that co-presence and shared interfaces provide complementary benefits: co-presence mainly supports workload, trust, and perceived partnership, while the shared interface mainly supports mutual awareness, coordination, and task performance.

This study has several limitations. The sample size of 16 participants (8 pairs) limits the generalisability of the findings, and the use of simulated robots with non-expert operators constrains direct applicability to real-world deployments. The controlled laboratory setting, while enabling fair comparison across conditions, does not fully capture the environmental uncertainty and time pressure of live disaster response operations.

Future work will focus on validating the interface with real robots and  domain experts in operational settings, and scaling to larger operator teams and more heterogeneous platforms. Using real-time physiological estimates to dynamically adapt the interface represents a key step toward fully adaptive HRI, and could be complemented by an assistive chatbot suggesting actions to operators under high workload. Exploring additional feedback modalities could further support operator awareness in demanding scenarios.

\section{Conclusion}
\label{sec:Conclusions}

This study designed, implemented, and evaluated a collaborative teleoperation interface for remotely operating multi-robot systems with a team of operators for critical tasks in challenging environments. A digital twin of a nuclear power plant was developed with simulated teleoperated mobile robots to complete disaster response scenarios requiring operator collaboration. A comparative study with 16 participants across three conditions varying interface sharing and operator co-location provided insights into task performance, team dynamics, cognitive load, and operator trust.

Co-located operators outperformed remote pairs on task completion time, cognitive load, and physiological stress indicators, highlighting the benefits of direct communication within the operator team. The unified shared interface further improved performance over individual interfaces across completion time, collision rate, and situational awareness accuracy, demonstrating the value of mutual situation awareness between operators.

These findings have broader implications for the design of human-robot systems in safety-critical domains, demonstrating that collaborative interface architectures can reduce miscommunication and improve situational awareness in hazardous environments. The results suggest that shared interfaces can meaningfully improve team dynamics, suggesting for their adoption in multi-robot operations such as disaster response and industrial inspection. Future work could explore how these benefits scale with larger operator teams, higher autonomy levels, or more complex tasks.

The collaborative teleoperation interface provides a practical solution for improving operator communication and mitigating the risk of errors due to miscommunication or lack of situation awareness in hazardous environments. This research demonstrates that such an interface can improve operator team dynamics and enable safer and more efficient multi-robot operations for complex tasks in challenging environments.

\section*{Acknowledgement}

This work was supported by the European Commission’s Marie Skłodowska-Curie Actions (MSCA) Project RAICAM (GA 101072634), and UK Research and Innovation (UKRI) grant number EP/X025977/1.

\balance

\bibliographystyle{IEEEtran}  
\bibliography{IEEEabrv}

\vspace{12pt}

\end{document}